\documentclass[journal]{IEEEtran}

\usepackage{cite}
\usepackage{hyperref}
\usepackage{cleveref}
\ifCLASSINFOpdf
\usepackage[pdftex]{graphicx}
\else
\fi

\usepackage{tikz}
\usetikzlibrary{arrows.meta}
\usepackage[utf8]{inputenc}
\usetikzlibrary{shapes.geometric, arrows}

\tikzset{%
  >={Latex[width=2mm,length=2mm]},
            base/.style = {rectangle, rounded corners, draw=black,
                           minimum width=4cm, minimum height=1cm,
                           text centered, font=\sffamily},
            process/.style = {base, minimum width=2.5cm, fill=none,
                           font=\ttfamily},
        }

\RequirePackage{luatex85}
\usepackage{pgfplots}
\pgfplotsset{compat=newest}
\pgfplotsset{every axis legend/.append style={%
cells={anchor=west}}
}
\usepgfplotslibrary{polar}
\usetikzlibrary{arrows}
\tikzset{>=stealth'}
\tikzstyle{startstop} = [rectangle, rounded corners, minimum width=3cm, minimum height=1cm,text centered, draw=black, fill=none]
\tikzstyle{action} = [trapezium, trapezium left angle=70, trapezium right angle=110, minimum width=3cm, minimum height=1cm, text centered, draw=black, fill=black!10]
\tikzstyle{process} = [rectangle, minimum width=3cm, minimum height=1cm, text centered, text width=3cm, draw=black, fill=none]
\tikzstyle{arrow} = [thick,->,>=stealth]

\definecolor{C1}{rgb}{0,    0,    0}
\definecolor{C2}{rgb}{0.4,    0.4,    0.4}
\definecolor{C3}{rgb}{0.6,    0.6,    0.6}
\definecolor{C4}{rgb}{0.,    0.,    0.}
\definecolor{C5}{rgb}{0.4,    0.4,    0.4}

\pgfplotsset{every axis legend/.append style={legend cell align=left}}

\usepackage{amsmath}
\usepackage{amsfonts}

\usepackage{siunitx}

\usepackage{algorithmic}

\usepackage{array}
\usepackage{booktabs}

\ifCLASSOPTIONcompsoc
 \usepackage[caption=false,font=normalsize,labelfont=sf,textfont=sf]{subfig}
\else
 \usepackage[caption=false,font=footnotesize]{subfig}
\fi

\title{Deep Reinforcement Learning for Event-Driven Multi-Agent Decision Processes}

\author{Kunal Menda, Yi-Chun Chen, Justin Grana, James W. Bono, Brendan D. Tracey, \\ Mykel J. Kochenderfer, David Wolpert
\thanks{K. Menda and M.J. Kochenderfer are with the Aeronautics and Astronautics Department at Stanford University, Stanford, CA USA (e-mail: \{kmenda,mykel\}@stanford.edu).

Y. Chen is with the Institute for Computational and Mathematical Engineering at Stanford University, Stanford, CA USA (email: yichunc@stanford.edu).

J. Grana is with the Santa Fe Institute, Santa Fe, NM USA (email: justin.grana@santafe.edu).

J.W. Bono is with Economists Incorporated, San Francisco, CA USA (email: jwbono@gmail.com).

B.D. Tracey is with the Santa Fe Institute, Santa Fe, NM USA, and is associated with the MIT Department of Aeronautics and Astronautics, Cambridge, MA USA (email: btracey@santafe.edu).

D. Wolpert is with the Santa Fe Institute, Santa Fe, NM USA, and is associated with the MIT Department of Aeronautics and Astronautics, Cambridge, MA USA, as well as Arizona State University, Tempe, AZ USA (email: david.h.wolpert@gmail.com).
}%

}

\begin{document}

\maketitle

\begin{abstract}

The incorporation of macro-actions (temporally extended actions) into multi-agent decision problems has the potential to address the curse of dimensionality associated with such decision problems. Since macro-actions last for stochastic durations, multiple agents executing decentralized policies in cooperative environments must act asynchronously. We present an algorithm that modifies Generalized Advantage Estimation for temporally extended actions, allowing a state-of-the-art policy optimization algorithm to optimize policies in Dec-POMDPs in which agents act asynchronously. We show that our algorithm is capable of learning optimal policies in two cooperative domains, one involving real-time bus holding control and one involving wildfire fighting with unmanned aircraft. Our algorithm works by framing problems as ``event-driven decision processes,'' which are scenarios where the sequence and timing of actions and events are random and governed by an underlying stochastic process. In addition to optimizing policies with continuous state and action spaces, our algorithm also facilitates the use of event-driven simulators, which do not require time to be discretized into time-steps. We demonstrate the benefit of using event-driven simulation in the context of multiple agents taking asynchronous actions. We show that fixed time-step simulation risks obfuscating the sequence in which closely-separated events occur, adversely affecting the policies learned. Additionally, we show that arbitrarily shrinking the time-step scales poorly with the number of agents.

\end{abstract}


\IEEEpeerreviewmaketitle

\section{Introduction}
\IEEEPARstart{I}{n} cooperative multi-agent environments, policies become difficult to optimize using reinforcement learning due to the curse of dimensionality. Additionally, the delay between accrued rewards and responsible action is a significant problem in cooperative multi-agent systems, where one agent may receive undeserved reward from favorable actions performed by other agents. A possible way to address these problems is to have agents learn multi-step \textit{macro-actions}, rather than the low-level \textit{primitive actions} typically considered in reinforcement learning algorithms. 

Suppose we were to use reinforcement learning to learn a policy to fly an unmanned aircraft to a nearby forest. Learning the precise actuator commands from the single reward of whether or not we got there would be intractable.  Macro-actions are typically defined by a policy that maps observations to primitive actions and a set of conditions specifying when the macro-action terminates \cite{sutton1999between}. We can pre-train macro-actions to perform higher level tasks such as way-point tracking, and have a policy learn to use them to navigate to our goal. 

When considering cooperative multi-agent environments, the problem of mapping a single reward for successful cooperation to each individual agent's primitive actions becomes exponentially more difficult as we scale the number of agents. For this reason, the use of macro-actions becomes substantially more necessary. For example, consider a problem in which many unmanned aircraft must cooperate to extinguish multiple fires. By restricting our action space to only high-level macro-actions (such as \textit{fly to a specified fire}, or \textit{attempt to extinguish a fire}), we can rely on traditional controllers to execute the low-level control policies required by these macro-actions. With macro-actions, we can substantially reduce the size of the state space (as many of the degrees-of-freedom pertaining to the pose of any given aircraft may not be relevant at this level of abstraction), as well as significantly reduce reward-delays and ease credit assignment. We can now attribute the rewards of extinguishing a large fire to the fact that we attempted to do so alongside a supporting agent, as opposed to a long sequence of low-level observations and actuator inputs.

Even when considering only \textit{decentralized} policies in which each agent makes an independent decision based on their partial observation of the environment, a challenge arises when extending the use of macro-actions to cooperative multi-agent domains. Since macro-actions are temporally extended, agents executing decentralized policies must act asynchronously. However, few off-the-shelf reinforcement learning algorithms consider the problem of multiple-agents acting asynchronously. To address asynchronicity, we view such a multi-agent decision process as event-driven, where agents choose a new macro-action when prompted by an event in some set of events occurring in the environment (such as completion of the currently executed macro-action, or the availability of new information). This article presents a method for learning optimal macro-action policies for event-driven decision processes. 

In many real world problems, the duration macro-actions are active are drawn from continuous distributions. However, current methods for planning and learning with macro-actions in multi-agent settings simulate the environment by discretizing time into fixed time-steps, as well as using discretized state and action spaces \cite{ghavamzadeh2006hierarchical}\cite{liu2016learning}\cite{shen2006multi}\cite{clement2016temporal}. When using macro-actions in multi-agent environments, temporal discretization poses a unique problem because assuming a fixed time step may group more than one temporally distinct event within a single time step. We refer to this as a \textit{race condition}. In these race conditions, information about the relative timing of the events is lost, and learning with such a simulation environment can lead to optimized policies that transfer poorly to the real world. 

As an example in which event-sequences can affect the policy learned, consider two aircraft which can each fly to one of two locations or remain in place. If they choose the same location to fly to, they both accrue a large negative reward (as they may collide and crash). If they choose different locations, they accrue a small positive reward. Given this, we may have that one agent chooses a location first, and the other observes that choice and decides to fly to the other location. However, if the time-step was large enough such that both agents always had to choose their action simultaneously, they may be forced to learn a strategy in which they both always choose the safe option of remaining in place. 

At the cost of computation time, one could mitigate the risk of race conditions by arbitrarily shrinking the time-step. However, we will demonstrate that in order to maintain a chosen threshold for the probability of a race condition, one must decrease the time-step quadratically with the number of agents. Since the computation time for a simulation is typically inversely related to the time-step, the computation time also scales quadratically with the number of agents. 

Representing policies as deep neural networks has shown success in the domain of decentralized multi-agent decision making \cite{guptacooperative}, and allows for the representation of continuous state-spaces when trained with policy gradient methods. This paper presents an extension of the PS-TRPO algorithm \cite{guptacooperative} to accommodate temporally extended actions, which both allows for continuous state-space representation and does not assume a fixed time-step. By using such an algorithm, we can simulate event-driven processes using event-driven simulators, which step from event to event as opposed to from time-step to time-step. By making use of such simulators, we can eliminate race conditions and scale the number of agents while only linearly scaling the computation time of the simulation. 

This article presents two key contributions. The first contribution is a modification to the PS-TRPO algorithm that allows it to optimize macro-action policies. In addition to interfacing with event-driven simulation and continuous state/action spaces, the algorithm does not require any expert demonstrations for policy optimization, unlike a state-of-the-art algorithm in this domain~\cite{liu2016learning}. We demonstrate that our algorithm is able to learn optimal policies in cooperative multi-agent environments, including real-time bus holding control and wildfire fighting with unmanned aircraft. The second contribution is a pair of experiments on the wildfire domain that demonstrate the utility of moving from fixed time-step simulations to event-driven simulations. The first of these experiments shows that large time-steps can result in race conditions that cause poor policy transfer between simulation and the real world, and the second demonstrates that mitigating the risk of race conditions forces a quadratic decrease in time-step with the number of agents. 

To our knowledge, in addition to being able to interface with event-driven simulators, our algorithm will be the first that uses deep reinforcement learning to optimize decentralized macro-action policies in multi-agent environments. While other research on macro-actions allow for planning over multiple levels of hierarchy, this paper will focus on the case in which learning is over macro-actions that exist at a single level of hierarchy, leaving the extension to full hierarchical learning to future work.

\section{Preliminaries and Related Work}

\subsection{Generalized Advantage Estimation and TRPO}

Policy gradient methods are widely used for optimizing policies through reinforcement learning. In these methods, we define a policy as a mapping from a history of observations $o_{0:k}$ to a distribution over actions at the $k$th decision-instant. A policy $\psi_\theta(o_{0:k})$ is parameterized using some parameter vector~$\theta$. Policy gradient methods update $\theta$ by estimating the gradient direction that improves performance the most.  Generalized Advantage Estimation (GAE) is a method for computing approximate policy gradients from simulation trajectories~\cite{schulman2015high}. Previous policy-gradient methods, though providing unbiased estimates of the policy gradient, result in gradient estimates with high variance that worsen with long time-horizons. Further, it is argued that GAE, which is a method specified by two hyper-parameters $\gamma$ and $\lambda$, reduces this variance while maintaining a tolerable level of bias~\cite{schulman2015high}. 

GAE defines the \textit{advantage function} $A(s_k,a_k)$ to be the difference between $V(s_k)$ and $Q(s_k, a_k)$, where $V(s_k)$ is the value being in state $s$ at the decision-instant $k$ and $Q(s_k,a_k)$ is the value of taking action $a$ from state $s$ at the decision-instant $k \in \mathbb{N}$. They incorporate the hyper-parameter $\gamma$ as follows:
 \begin{align}
 V^{\gamma}(s_k) &:= \mathbb{E}_{s_{k+1:\infty}, a_{k:\infty}} \left[\sum_{l=0}^\infty \gamma^l r_{k+l}\right] \\
 Q^{\gamma}(s_k,a_k) &:= \mathbb{E}_{s_{k+1:\infty}, a_{k+1:\infty}} \left[\sum_{l=0}^\infty \gamma^l r_{k+l}\right] \\ 
 A^{\gamma}(s_k,a_k) &:= Q^{\gamma}(s_k,a_k) - V^{\gamma}(s_k) 
 \end{align}
where $r_{k}$ is the reward received from taking action $a_k$ from state $s_k$. The discounted approximation to the policy-gradient is then defined as:
\begin{equation}
 g^\gamma := \mathbb{E}_{s_{0:\infty}, a_{0:\infty}} \left[ \sum_{k=0}^\infty A^{\gamma}(s_k, a_k) \nabla_\theta \log \psi_\theta (a_k\mid s_k) \right]
\end{equation}

When a policy is used to simulate a batch of episodes in one training epoch, the GAE advantages are estimated by first computing $\delta^V_k$, which is an estimate of the immediate advantage of action of $a_k$, and then the GAE advantage estimate $A^{GAE(\gamma,\lambda)}_k$ as follows:
\begin{align}
 \delta^V_k :&= r_k + \gamma V(s_{k+1}) - V(s_k) \label{eqn:gae_delta}\\
 A^{GAE(\gamma,\lambda)}_k &= \sum_{l=0}^\infty (\gamma\lambda)^l\delta^V_{k+l} \label{eqn:gae_adv}
\end{align}
Here, $V(s_k)$ is the approximate value function, referred to as a \textit{baseline}, and $\lambda$ is a hyper-parameter that is used in addition to $\gamma$ to control the bias-variance trade-off in gradient estimation. The baseline $V(s_k)$ is also a function approximator (such as a linear mapping or neural network) parameterized by some vector $\phi$. Once the advantages are computed, an algorithm such as Trust-Region Policy Optimization (TRPO) updates the parameters $\theta$ by constraining the KL-divergence between the previous and new function approximations, and updates $\phi$ by training the baseline model with supervised learning to map states to their average discounted returns in the set of trajectories \cite{schulman2015trust}.

\subsection{Dec-POMDPs and PS-TRPO}

Initial efforts to extend the partially observable Markov decision process (POMDP) framework to multi-agent settings attempted centralized control over the joint state and action spaces of all agents, making planning intractable. To address this problem, efforts have been dedicated to solving \textit{decentralized} versions of the same problems, in which each agent has access to only some local observation of the state space (which may include additional information communicated by other agents), and must choose their action based on only that observation. 
PS-TRPO is an extension to the TRPO algorithm to cooperative multi-agent domains modeled as decentralized POMDPs (Dec-POMDPs), in which multiple agents act in a single environment with decentralized execution of the same policy $\psi_\theta$~\cite{guptacooperative}. Here agent $i$'s $k$th action is given by $a_{i,k} \sim \psi_\theta(o_{i,0:k})$, where $o_{i,0:k}$ is the $i$th agent's observation history. The parameters $\theta$ of $\psi_\theta$ are then updated by the TRPO algorithm using advantages computed from all agents' trajectories. They assume a reward structure in which all rewards are shared jointly by agents.

\subsection{MacDec-POMDPs and The PoEM Algorithm}
\label{sec:macdecpomdps}

A policy in a Dec-POMDP is a mapping from an agent's local observations to their local action. However, we may want to optimize a policy in an environment in which the observation spaces high-dimensional. 
There often exists a natural hierarchical abstraction over the policy space, in which high-level controllers issue commands to low-level controllers, which lead to the development of the \textit{options} framework \cite{sutton1999between}. An agent $i$'s option, or macro-action space $M_i$ is defined as containing macro-action tuples $m_i = \langle I_i^m, \beta_i^m, \pi_i^m \rangle$, where for each macro-action $m_i \in M_i$, $I_i^m$ specifies a set of states from which the macro-action can be initiated, $\beta_i^m$ specifies the probability of the action terminating in any given state after having been initiated, and $\pi_i^m$ specifies the low-level controller corresponding to that macro-action, mapping the agent's observation to a primitive (non-macro) action.

Planning over macro-actions in Dec-POMDPs can be formalized as a \textit{MacDec-POMDP} \cite{amato2014planning}. Since macro-actions can extend over arbitrary time horizons, we must treat our decision process as a semi-MDP and find an optimal policy $\psi$ that maps an agent's observation history to a macro-action. The current state-of-the-art model-free reinforcement learning algorithm for optimizing policies in the MacDec-POMDP setting is the PoEM algorithm \cite{liu2016learning}. This algorithm aims to optimize a finite state controller (FSC) that represents a macro-action policy using expectation-maximization (EM). 
The algorithm is shown to scale linearly with the number of agents, but quadratically with the number of \textit{nodes} in the FSC, a number which is proportional to the discretization of the observation and action space. Additionally, the algorithm generates experience histories by requiring demonstrations from an expert heuristic controller. In this paper, we will extend the PS-TRPO algorithm to optimize policies in MacDec-POMDPs, preserving the algorithms' ability to optimize over continuous state and action spaces and requiring no expert demonstrations for policy optimization.  

\section{Extending PS-TRPO To MacDec-POMDPs}
To extend the PS-TRPO algorithm to optimize macro-action policies, we must address the fact that actions are asynchronous and temporally extended. We begin by framing a MacDec-POMDP as an event-driven process, allowing us to associate decision instants to discrete events. With this framework, we can modify GAE to allow PS-TRPO to optimize a policy with trajectories generated from MacDec-POMDPs.

\subsection{MacDec-POMDPs as Event-Driven Processes}

Similar to the definition of the option presented in Section \ref{sec:macdecpomdps}, we define the macro-action space as the tuple $\langle I_i^m, E_i^m, \pi_i^m \rangle$. For each macro-action $m$, $I_i^m$ and $\pi_i^m$ are the set of states from which agent $i$ can take the macro-action and the lower-level controller the macro-action specifies, respectively. However, $E_i^m$ now specifies a set of events, the triggering of any of which terminates the macro-action, prompting the agent to select another macro-action given a new observation. In our example where unmanned aircraft coordinate to extinguish fires, the macro-action chosen by one aircraft may be to fly to a particular fire. A lower-level policy tuned to optimally carry out this order would take over control until a relevant event occurs, such as receiving information that the fire being flown to was just extinguished. When such an event occurs, the agent would be given a new observation and the opportunity to choose a new action. 

Unlike conventional MDPs, agents may receive rewards at any instant in time, not just when an event prompts them to act. For example, if rewards were given to the whole team for accomplishments, one agent may accrue a reward for the team without triggering an event that prompts the another agent to choose a new action. Hence, we adopt a convention by which all rewards accrued by an agent over the duration $\Delta t_{i,k}$, in which macro-action $m_{i,k}$ is being carried out, can be collected into a single reward to associate with that macro-action. We let:
\begin{align}
r_{i,k} &= \sum_{j=0}^{J_k} e^{-\gamma\Delta \tau_j}\rho_j + \int_{0}^{\Delta t_{i,k}}e^{-\gamma t} c_{i,k}(T_{i,k}+t) dt \\
T_{i,k} &= \sum_{j=0}^{k-1}\Delta t_{i,j}
\end{align}
Here, $\rho_{0:J_k}$ are the $J_k$ discrete rewards accrued by agent $i$ at times $T_{i,k} + \Delta \tau_{0:J_k}$, $c_{i,k}(t)$ is a continuous-time reward generator over the interval that the macro-action is being executed, and $\gamma$ is the continuous-time discount rate. The rewards $\rho_{0:J_k}$ are accrued on events occurring in the environment during the time when an agent's $k$th macro-action is active.

\subsection{Adapting GAE for Temporally Extended Actions}
A simulation episode will now generate a trajectory $\langle\ldots,o_{i,k},m_{i,k},\Delta t_{i,k}, r_{i,k},\ldots\rangle$ for each agent $i$. With these trajectories, we can modify Equations \ref{eqn:gae_delta} and \ref{eqn:gae_adv} in GAE to account for the temporal extension of actions. We will refer to GAE modified for macro-actions as M-GAE. The modified equations are:
\begin{align}
  \delta^V_{i,k} &= r_{i,k} + e^{-\gamma\Delta t_{i,k}} V(o_{i,k+1}) - V(o_{i,k}) \\
  A^{M-GAE(\gamma,\lambda)}_{i,k} &= \sum_{l=0}^\infty e^{-\gamma\lambda (T_{i,l} - T_{i,k})}\delta^V_{k+l}
\end{align}
The modifications now factor in the duration over which macro-actions are active into the discounted sum of rewards that is associated to the taking of each macro-action, as opposed to assuming every action lasts a single time-step.

Using M-GAE, the PS-TRPO algorithm can now optimize policies in MacDec-POMDPs framed as event-driven processes.

\section{Event-Driven Simulators}

Many methods for learning policies in multi-agent settings assume that simulations are performed with some fixed step-size. However, in reality, the timings of the various agents' decisions may be drawn from continuous, stochastic distributions. As mentioned earlier, using simulators with fixed time-steps leaves the possibility open that more than one relevant event can occur in the same time-step. Consequently, a fixed time-step simulation obfuscates the relative timing of these events, and may result in policies being learned on the simulator that transfer poorly to the real world, where the sequences may be important.

A key advantage of framing MacDec-POMDPs as event-driven processes is that it allows us to use event-driven simulation. Event-driven simulators, such as the Python module \textit{SimPy} \cite{simpy}, step the environment from event to event as opposed to from one time-step to the next. Underlying the simulation is the assumption that all entities cycle through first performing some instantaneous processing, and then \textit{yielding} until some event they specify occurs, which could be the lapse of some specified amount of time. 

Even though not every real-world environment is event-driven, modeling one as an event-driven process does not necessarily sacrifice much simulation fidelity. For example, suppose that we have two aircraft, intending to fly to two different fires. We can simulate their individual flights by sampling a duration for each aircraft from a distribution dependent on their desired flight path and average velocities. We can then assert that after those durations have elapsed, each aircraft will have reached their respective fires. By doing so, we can treat the initiation and termination of each agent's flight as distinct events, allowing the simulator to skip simulating anything in-between. In this example, we do, however, lose the ability to simulate the influence flight paths have on each other. This simplification may be benign because we want to learn high-level policies that may not need to take into account these details.

To motivate the use of event-driven simulation, in the latter part of the next section we will demonstrate that the choice of step-size can have adverse affects on the quality of policy learned in simulation. Though we can always shrink the step-size by a sufficient amount to ensure high-fidelity simulation, we will show through a set of Monte-Carlo experiments that doing so causes the computational cost of simulation to scale poorly with the number of agents compared to an event-driven simulation.

\section{Performance Experiments}

The experiments in this section validate the claim that the PS-TRPO algorithm with M-GAE is able to optimize policies in multi-agent environments where agents select from a space of temporally extended actions. The real-time bus holding problem will be used to validate this claim. This section also shows the benefit of using event-driven simulators over fixed time-step simulators in the wildfire fighting problem.

\subsection{Real-Time Bus Holding}
\label{sec:bbcontrol}

\begin{figure}
  \centering
  \scalebox{0.6}{\input{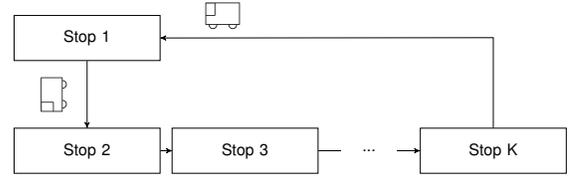}}
  \caption{An example bus corridor. \label{fig:buscorridor}}
\end{figure}

Consider a fleet of buses servicing the bus-corridor, shown in \Cref{fig:buscorridor}. Buses sequentially visit a set of stops numbered $1$ though $K$, returning to Stop 1 after the loop completes, unloading all passengers, and repeating the cycle. A phenomenon called \textit{bus bunching} can occur, in which buses queue to visit a set of stops, leaving no time between their arrivals, after which a long delay occurs until the bunch cycles back to the stop. This phenomenon leads to unreliable and sub-optimal service, and the research community has explored a strategy called \textit{bus holding} to mitigate the issue. In this control strategy, a bus is instructed to \textit{hold} at its current stop for some specified amount of time, in addition to the time it waits for passenger boarding and alighting. The problem has previously been framed as a Dec-POMDP, and effective decentralized policies have been discovered through reinforcement learning~\cite{chen2016real}. We will follow the problem formulation described by Chen et al. \cite{chen2016real}, and show that PS-TRPO with M-GAE generates polices that match their results in performance.

\subsubsection{Problem Specification}
Our environment consists of a bus corridor of $N$ buses and $K$ bus stops. All agents are initialized in a queue for access to Stop 1, with Bus 1 being the first in the queue and Bus $N$ being the last. Upon arrival at Stop $k$, $U_{i,k} = \lfloor q_k\cdot L_{i,k} \rceil$ passengers will attempt to alight the bus, where $q_k \in [0,1]$ and $L_{i,k} \in [0,L_\text{max}]$ is the load of the bus when reaching Stop~$k$. Here, $\lfloor \cdot \rceil$ is used to indicate rounding to the nearest integer. Additionally, $B_{i,k} = \min(\lfloor \nu_k\cdot h_{i,k} \rceil, L_\text{max} - L_{i,k} + U_{i,k})$ passengers will attempt to board the bus, where $\nu_k \in [0,\infty)$ is the passenger arrival rate, and $h_{i,k}$ is the \textit{headway} between Bus~$i$ and Bus~$i-1$ ahead of it, defined as the time elapsed between the departure of Bus~$i-1$ and the arrival of Bus~$i$ at Stop~$k$. The bus will wait a nominal amount of time of $S_{i,k} = \max(t_a U_{i,k}, t_b B_{i,k})$, where $t_a$ and $t_b$ are the rate at which passengers alight and board, respectively. The bus will then wait an additional amount of time chosen from its action space before departing from the stop. The bus will then travel for time $rt_{k+1}$ to Stop~$k+1$, and join the queue for arrival at that stop.

Upon arriving at a stop, Bus $i$ will receive an observation of $[z_1, z_2, z_3, z_4]$, where $z_1$ is the current stop's index, $z_2$ is the headway $h_{i,k}$, $z_3$ is the current load $L_{i,k}$, and $z_4$ is the sojourn-time elapsed between its current and previous observations. The bus then selects a holding time as some multiple $a_{i,k} \in \{0,1,2,3\}$ of the parameter $T_\text{hold}$. The goal of policy optimization is for each bus to choose actions in such a manner that all headways $h_{i,k}$ are as close as possible to the planned headway $H$. Hence, whenever a bus $i$ arrives at a stop $k$, a reward of $-\nu_k |h_{i,k} - H|^2$ will be accrued by all agents.

The parameters for the bus corridor environment used in our experiments are summarized in \Cref{tab:bbparams}. All policies are trained on maximum episode lengths of three simulated hours, which corresponds to four to five cycles of all buses through the corridor. The policy trained using PS-TRPO with M-GAE is represented by a Multi-Layer Perceptron (MLP). The parameters to represent the policy and train it using PS-TRPO with M-GAE are summarized in \Cref{tab:bbparams_pstrpo}. 

As benchmarks for comparison, we compare the trained MLP policy to the performance of not holding at all, which we refer to as \textit{No Holding}, and a policy that we call \textit{Optimized Thresholds}. In this policy, a set of three ordered thresholds $T_1 > T_2 > T_3$ are given, and the action chosen is:
\begin{equation}
  a_{i,k}(h_{i,k})=\begin{cases} 
    0 & \text{if } h_{i,k} > T_1 \\
    1 & \text{if } T_2 < h_{i,k} < T_1 \\
    2 & \text{if } T_3 < h_{i,k} < T_2 \\
    3 & \text{if } h_{i,k} < T_3 
  \end{cases}
\end{equation}

The values $T_{1:3}$ are optimized using the SciPy implementation of differential evolution~\cite{storn1997differential}\cite{scipy}. The optimization target was average return under the same reward function used to optimize the MLP policy with the same episode length.

\begin{table}
\centering
\caption{Environment parameters used in Real-Time Bus Holding experiments}
\label{tab:bbparams}
\begin{tabular}{@{}llr@{}} \toprule
Parameter & Value & Unit \\ \midrule
$N$ & 6 & Buses \\
$K$ & 10 & Stops \\
$L_\text{max}$ & 75 & pax \\
$rt_{k=\{1,..,10\}}$ & 180 & \si{\second}\\
$t_a$ & 1.8 & \si{\second}  \\
$t_b$ & 3.0 & \si{\second}  \\
$ H $ & 6.0 & \si{\minute} \\
$ T_\text{hold}$ & 30.0 & \si{\second} \\
\midrule
Stop & $q_k$ & $\nu_k~\left[\frac{\text{pax}}{\si{\minute}}\right]$ \\
\midrule
1 & 1.0 & 1.5 \\
2 & 0 & 2.25  \\
3 & 0.1 & 1.4 \\ 
4 & 0.25 & 4.5 \\
5 & 0.25 & 2.55 \\
6 & 0.5 & 1.8 \\
7 & 0.5 & 1.43 \\
8 & 0.1 & 1.05 \\
9 & 0.75 & 0.75 \\
10 & 0.1 & 0.45 \\

\bottomrule
\end{tabular}
\end{table}

\begin{table}
\centering
\caption{Parameters used for MLP policy representation and training in Real-Time Bus Holding experiments}
\label{tab:bbparams_pstrpo}
\begin{tabular}{@{}llr@{}} \toprule
Parameter & Value & Unit \\ \midrule
MLP Hidden Size & 32 & neurons \\
$\gamma$ & $10^{-5}$ & \si{\per\second}\\
$\lambda$ & 2.0 & \\
TRPO Max Step & 0.01 & \\
Batch Size & 540 & $\frac{\text{episodes}}{\text{epoch}}$ \\
\bottomrule
\end{tabular}
\end{table}

\subsubsection{Results}

\begin{figure}
  \centering
  \scalebox{0.9}{\input{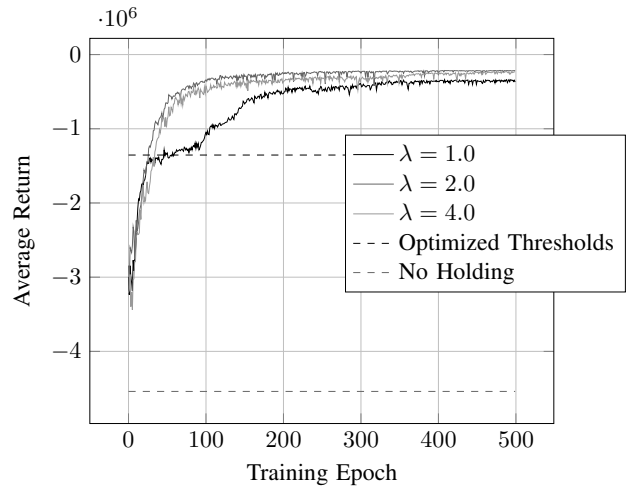}}
  \caption{Training curves for MLP policy on the Real-Time Bus Holding Environment, for various values of the parameter $\lambda$.}
  \label{fig:bbtraining}
\end{figure}

\begin{figure}
  \centering
  \scalebox{0.9}{\input{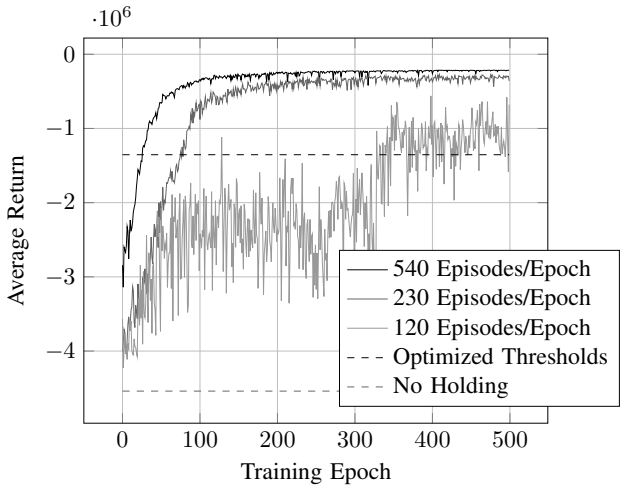}}
  \caption{Training curves for MLP policy on the Real-Time Bus Holding Environment, trained with various batch sizes.}
  \label{fig:bbtest}
\end{figure}

\begin{figure*}
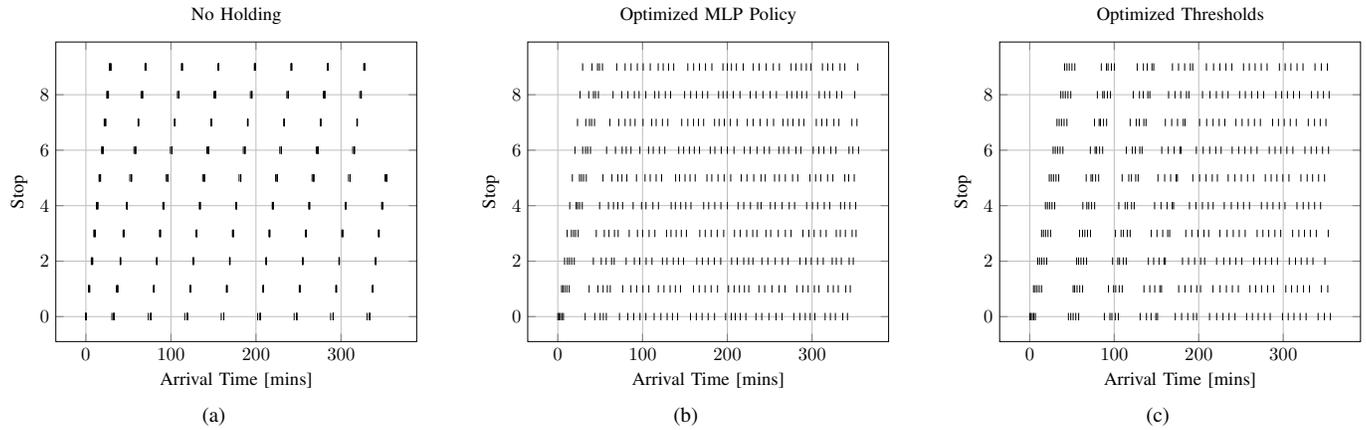
%
\centering
\subfloat[][\label{fig:bbnoholdarriv}]{\scalebox{0.7}{\input{nohold_arriv}}}%
\qquad
\subfloat[][\label{fig:bbrlpolarriv}]{\scalebox{0.7}{\input{rlpol_arriv}}}
\qquad
\subfloat[][\label{fig:bboptimarriv}]{\scalebox{0.7}{\input{optim_arriv}}}
\caption{Arrival times of the six buses at the ten stops, under (a) a policy that uses no holding, (b) a policy learned via PS-TRPO using M-GAE, and (c) a policy that optimizes three headway thresholds for selecting between the actions.}%
\label{fig:bbarrival_times}%
\end{figure*}

\begin{figure*}%
\centering
\subfloat[][\label{fig:bbnoholdload}]{\scalebox{0.68}{\begin{tikzpicture}[]
\begin{axis}[title = {No Holding}, xlabel = {Stop}, grid=both, ylabel = {Load at Arrival [pax]}]\addplot+ [mark = {o}, black, only marks = {true}]coordinates {
(9, 50)
(0, 64)
(1, 60)
(2, 75)
(3, 75)
(4, 75)
(5, 75)
(6, 75)
(7, 75)
(8, 75)
(9, 50)
};
\addplot+ [mark = {o}, black, only marks = {true}]coordinates {
(9, 0)
(0, 0)
(1, 0)
(2, 0)
(3, 0)
(4, 0)
(5, 0)
(6, 0)
(7, 0)
(8, 0)
(9, 0)
};
\addplot+ [mark = {o}, black, only marks = {true}]coordinates {
(9, 0)
(0, 0)
(1, 0)
(2, 0)
(3, 0)
(4, 0)
(5, 0)
(6, 0)
(7, 0)
(8, 0)
(9, 0)
};
\addplot+ [mark = {o}, black, only marks = {true}]coordinates {
(9, 0)
(0, 0)
(1, 0)
(2, 0)
(3, 0)
(4, 0)
(5, 0)
(6, 0)
(7, 0)
(8, 0)
(9, 0)
};
\addplot+ [mark = {o}, black, only marks = {true}]coordinates {
(9, 0)
(0, 0)
(1, 0)
(2, 0)
(3, 0)
(4, 0)
(5, 0)
(6, 0)
(7, 0)
(8, 0)
(9, 0)
};
\end{axis}

\end{tikzpicture}}}%
\qquad
\subfloat[][\label{fig:bbrlpolload}]{\scalebox{0.68}{\begin{tikzpicture}[]
\begin{axis}[title = {Optimized MLP Policy}, xlabel = {Stop}, grid=both, ylabel = {Load at Arrival [pax]}]\addplot+ [mark = {o}, black, only marks = {true}]coordinates {
(9, 13)
(0, 16)
(1, 11)
(2, 28)
(3, 36)
(4, 59)
(5, 61)
(6, 40)
(7, 26)
(8, 28)
(9, 10)
};
\addplot+ [mark = {o}, black, only marks = {true}]coordinates {
(9, 11)
(0, 13)
(1, 9)
(2, 22)
(3, 28)
(4, 43)
(5, 45)
(6, 32)
(7, 21)
(8, 23)
(9, 8)
};
\addplot+ [mark = {o}, black, only marks = {true}]coordinates {
(9, 12)
(0, 14)
(1, 10)
(2, 24)
(3, 31)
(4, 49)
(5, 54)
(6, 39)
(7, 27)
(8, 29)
(9, 9)
};
\addplot+ [mark = {o}, black, only marks = {true}]coordinates {
(9, 11)
(0, 13)
(1, 10)
(2, 25)
(3, 32)
(4, 51)
(5, 54)
(6, 39)
(7, 28)
(8, 32)
(9, 12)
};
\addplot+ [mark = {o}, black, only marks = {true}]coordinates {
(9, 10)
(0, 12)
(1, 11)
(2, 26)
(3, 33)
(4, 53)
(5, 58)
(6, 42)
(7, 31)
(8, 36)
(9, 14)
};
\end{axis}

\end{tikzpicture}}}
\qquad
\subfloat[][\label{fig:bboptimload}]{\scalebox{0.68}{\begin{tikzpicture}[]
\begin{axis}[title = {Optimized Thresholds}, xlabel = {Stop}, grid=both, ylabel = {Load at Arrival [pax]}]\addplot+ [mark = {o}, black, only marks = {true}]coordinates {
(9, 29)
(0, 33)
(1, 21)
(2, 51)
(3, 64)
(4, 75)
(5, 75)
(6, 56)
(7, 43)
(8, 49)
(9, 18)
};
\addplot+ [mark = {o}, black, only marks = {true}]coordinates {
(9, 12)
(0, 14)
(1, 10)
(2, 23)
(3, 30)
(4, 48)
(5, 53)
(6, 39)
(7, 29)
(8, 34)
(9, 13)
};
\addplot+ [mark = {o}, black, only marks = {true}]coordinates {
(9, 11)
(0, 13)
(1, 9)
(2, 23)
(3, 30)
(4, 47)
(5, 51)
(6, 37)
(7, 28)
(8, 32)
(9, 13)
};
\addplot+ [mark = {o}, black, only marks = {true}]coordinates {
(9, 10)
(0, 12)
(1, 9)
(2, 21)
(3, 28)
(4, 43)
(5, 46)
(6, 34)
(7, 26)
(8, 30)
(9, 12)
};
\addplot+ [mark = {o}, black, only marks = {true}]coordinates {
(9, 1)
(0, 2)
(1, 4)
(2, 11)
(3, 15)
(4, 27)
(5, 30)
(6, 24)
(7, 20)
(8, 24)
(9, 10)
};
\end{axis}

\end{tikzpicture}}}
\caption{Load at arrival of the six buses as they arrive at the 10 stops, under (a) a policy that uses no holding, (b) a policy learned via PS-TRPO using M-GAE, and (c) a policy that optimizes three headway thresholds for selecting between the actions. Results are displayed for the buses' fourth cycle through the stops to show the steady-state behavior.}%
\label{fig:bbload}%
\end{figure*}
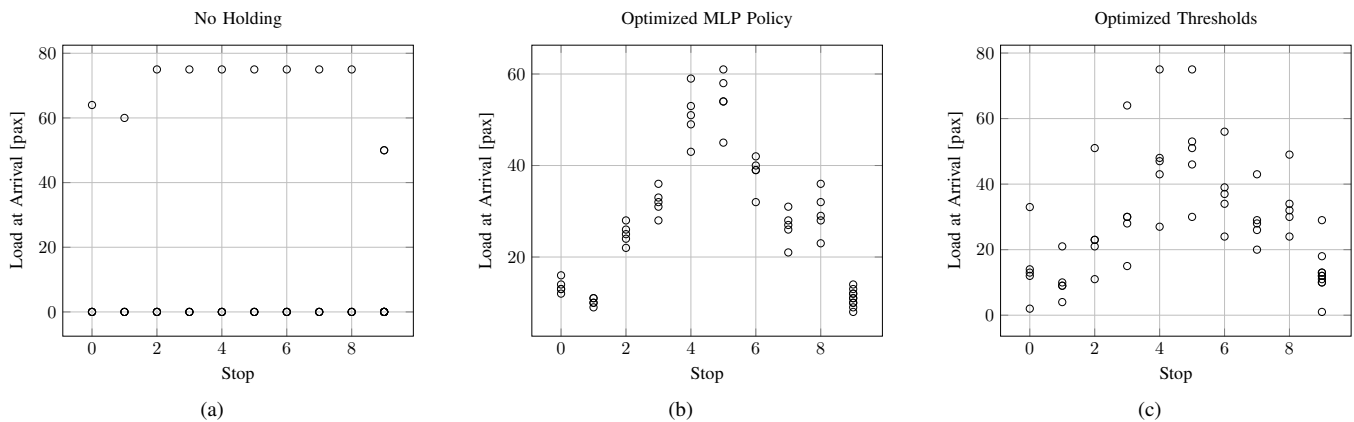

As shown in \Cref{fig:bbtraining,fig:bbtest}, the optimized MLP policy is able to achieve an average total return per episode that is substantially greater than those of the No Holding and Optimized Thresholds policies, though the Optimized Thresholds policy is itself a substantial improvement over No Holding. In \Cref{fig:bbtraining}, we see that $\lambda$ slightly influences the total return, but the tested values produce policies that are relatively similar in performance when compared to the Optimized Thresholds. In \Cref{fig:bbtest}, we do, however, observe that decreasing the batch size degrades convergence. This degradation is due to the fact that the estimated gradients are noisier with smaller batches. 

\Cref{fig:bbarrival_times,fig:bbload} show that the No Holding policy results in the fleet of buses moving together as a single bunch, as expected. The entire load is borne by the leading bus, which saturates to being full, and the buses behind it remain almost empty. On the other hand, the Optimized Thresholds policy has similar steady-state performance as the optimized MLP policy, but takes approximately 200 minutes from the start of the episode to achieve uniformity in bus arrivals at stops, while the optimized MLP policy requires only around 100 minutes. Neither of these two policies saturate the load of any bus in steady-state operation, but the optimized MLP policy maintains lower variance in the distribution of loads over buses when they reach any given stop. Though the variance in the distribution of loads over buses is not explicitly optimized for, the variance occurs as a consequence of non-uniformity in arrival times. Hence, smaller variance in this distribution of loads suggests that the optimized MLP policy is mitigating the adverse effects of bus-bunching more effectively than Optimized Thresholds can. This experiment provides evidence that the PS-TRPO algorithm using M-GAE is well suited to optimize policies with continuous observation spaces in event-driven, multi-agent environments. 

\subsection{Wildfire Fighting Problem}
\label{sec:feprob}

The goal of this experiment is to demonstrate the utility of using event-driven simulations, which step from event-to-event, over fixed time-step simulations. We introduce a wildfire fighting environment. We will learn policies on both an event-driven simulation of this environment, which makes no assumption regarding time-step, as well as fixed time-step simulations of the environment, where events that occur within the same time-step are considered to have occurred simultaneously. We will then test all learned polices on the event-driven simulation to examine whether approximation errors introduced by temporal discretization can degrade the performance of the policy in the event-driven environment. As the time-step goes to zero, the simulation should be effectively identical to the event-driven simulation. Thus, it is expected that a sufficiently small time-step should lead to minimal to no degradation in the performance of the policy, when tested on the event-driven simulator.

\subsubsection{Problem Specification}

There are $N$ unmanned aircraft that are tasked with extinguishing $K$ fires. At a decision-instant, an aircraft can choose any of the five closest fires to move straight toward at velocity $v_\text{uav}$, or to hold its current location for a time $T_\text{hold}$. If an aircraft is currently at a fire, and it chooses to hold, it will attempt to extinguish the fire. Each fire is given a \textit{health} $T_\text{health}$, which is the amount of time aircraft must collectively attempt to extinguish the fire before it is extinguished. When any fire is extinguished, all agents receive a reward of $r_\text{ext}$. If an aircraft is moving toward a fire or currently attacking it, it is considered \textit{interested} in the fire. However, if an aircraft attempts to extinguish a fire that another aircraft is currently attempting to extinguish, all agents receive a penalty of $r_\text{pen}$ at the instant it makes that decision. 

At any decision instant $k$, aircraft $i$ receives an observation $o_{i,k}$ consisting of its own $x$ and $y$ location, its distance to the five closest fires, the interest in each of those fires, whether each of those fires has already been extinguished, and the remaining health of each of the fires. A simulation episode lasts a maximum simulated duration of $T_\text{max}$. The parameters used by our simulation environment are summarized in~\Cref{tab:feparams}. The nine fires form three clusters of three, with the cluster centroids evenly spaced 0.99 \si{\meter} from the origin, and with the fires evenly spaced 0.01 \si{\meter} from their centroids. The initial locations of the aircraft are randomized uniformly within the bounds of $[-1,1]$ \si{\meter} at the start of each episode. Once again, an MLP is used to represent the decentralized policy. The parameters used to train policies using PS-TRPO with M-GAE are summarized in~\Cref{tab:feparams_pstrpo}.

\begin{table}
\centering
\caption{Environment parameters used in the wildfire fighting problem experiments}
\label{tab:feparams}
\begin{tabular}{@{}llr@{}} \toprule
Parameter & Value ($\pm \sigma$) & Unit \\ \midrule
$N$ & 3 & Aircraft \\
$K$ & 9 & Fires \\
$v_\text{uav}$ & 0.015 & $\si[per-mode=symbol]
{\metre\per\second}$ \\
$T_\text{hold}$ & 3.0 ($\pm 0.3$) & \si{\second}\\
$T_\text{health}$ & 3.0 & \si{\second}  \\
$T_\text{max}$ & 265 & \si{\second} \\
$ r_\text{ext} $ & 1.0 & points \\
$ r_\text{pen}$ & 20.0 & points \\
\bottomrule
\end{tabular}
\end{table}

\begin{table}
\centering
\caption{Parameters used for MLP policy representation and training in the wildfire fighting problem experiments}
\label{tab:feparams_pstrpo}
\begin{tabular}{@{}llr@{}} \toprule
Parameter & Value & Unit \\ \midrule
MLP Hidden Size & 32 & neurons \\
$\gamma$ & $0.02$ & \si{\per\second}\\
$\lambda$ & 1.0 & \\
TRPO Max Step & 0.01 & \\
Batch Size & 600$\sim$1200 & $\frac{\text{episodes}}{\text{epoch}}$ \\
\bottomrule
\end{tabular}
\end{table}

\begin{figure}
  \centering
  \scalebox{0.7}{



\begin{tikzpicture}[node distance=2cm]

\node (atlivefire) [process] {At live fire?};
\node (nootherinterest) [process, below of=atlivefire] {Other interest?};
\node (noidec) [startstop, below of=nootherinterest] {$p\sim \mathcal{U}(0,1) < 0.84$?};
\node (ioidec) [startstop, right of=nootherinterest, xshift=2.5cm] {$p\sim \mathcal{U}(0,1) < 0.27$?};
\node (hold) [action, below of=ioidec] {Hold};
\node (probs)[rectangle, below of=noidec, yshift=-2cm, xshift=2cm] {\scalebox{0.7}{\input{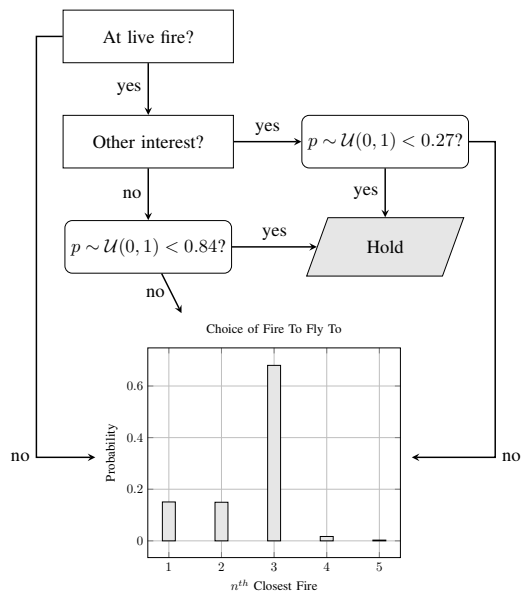}}};
\node (dummy) [left of=atlivefire] {};

\draw [arrow] (atlivefire) -- node[anchor=east] {yes} (nootherinterest);
\draw [arrow] (nootherinterest) -- node[anchor=east] {no} (noidec);
\draw [arrow] (nootherinterest) -- node[anchor=south] {yes} (ioidec);
\draw [arrow] (ioidec) -- node[anchor=east] {yes} (hold);
\draw [arrow] (noidec) -- node[anchor=south] {yes} (hold);
\draw [arrow] (noidec) -- node[anchor=east] {no} (probs);
\draw [arrow] (atlivefire.west) -- ++(-0.5,0) |- node[anchor=east] {no} (probs.west);
\draw [arrow] (ioidec.east) -- ++(0.5,0) |- node[anchor=west] {no} (probs.east);

\end{tikzpicture}

}
  \caption{Approximated behavior of the learned MLP policy in the wildfire fighting problem.}
  \label{fig:appxpolicy}
\end{figure}

\subsubsection{Results}

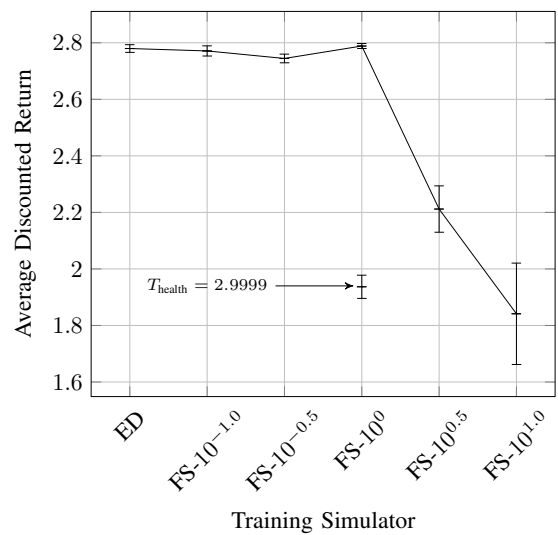
\begin{figure}
  \centering
  \scalebox{0.9}{\begin{tikzpicture}[]
\begin{axis}[xlabel = {Training Simulator}, grid=both, ylabel = {Average Discounted Return},xtick={1,2,3,4,5,6}, xticklabels={ED, FS-$10^{-1.0}$, FS-$10^{-0.5}$, FS-$10^{0}$, FS-$10^{0.5}$, FS-$10^{1.0}$},xticklabel style={rotate=45}]\addplot+ [
mark = {black}, black, error bars/.cd, 
x dir=both, x explicit, y dir=both, y explicit]
table [
x error plus=ex+, x error minus=ex-, y error plus=ey+, y error minus=ey-
] {
x y ex+ ex- ey+ ey-
4.0 1.93681589969 0.0 0.0 0.0410598352529 0.0410598352529
};\addplot+ [
mark = {black}, black, error bars/.cd, 
x dir=both, x explicit, y dir=both, y explicit]
table [
x error plus=ex+, x error minus=ex-, y error plus=ey+, y error minus=ey-
] {
x y ex+ ex- ey+ ey-
1.0 2.77967987039 0.0 0.0 0.0136182751054 0.0136182751054
2.0 2.77116278061 0.0 0.0 0.0179833271993 0.0179833271993
3.0 2.74452130859 0.0 0.0 0.0153417433698 0.0153417433698
4.0 2.78854871988 0.0 0.0 0.00882454950388 0.00882454950388
5.0 2.21184437795 0.0 0.0 0.0820432066556 0.0820432066556
6.0 1.8413277904 0.0 0.0 0.179313897733 0.179313897733
};
\node (s) [rectangle] at (axis cs:2, 1.94) {\scriptsize{$T_\text{health} = 2.9999$}};
\node (e) at (axis cs:4, 1.94) {};
\draw[->] (s) -- (e);
\end{axis}

\end{tikzpicture} }
  \caption{Performance when tested on the Event-Driven (ED) simulator of policies learned on the (ED) simulator and Fixed Step-size (FS) simulations of various step sizes. Performance drops substantially when the FS-$10^0$ simulator has $T_\text{health} = 2.9999$ instead of $T_\text{health} = 3.0$. }
  \label{fig:transferperformance}
\end{figure}

To ensure that intelligent policies can be learned in this problem domain, we first trained a policy on the event-driven simulation for an excessive 3000 epochs. The learned behavior was then examined to see if actions taken by the policy match intuition. Since there is a large penalty for attempting to hold at a fire that another agent is currently holding at, intuition would suggest that doing so when we observe another agent's interest in the fire should be avoided. Meanwhile, it would be sensible to hold at a fire if it does not have another agent interested in it. We find that agents only hold at a fire that has another agent interested in it 27\% of the time, while they hold at a fire that has no other agent interested in it 84\% of the time. If the agent chooses another fire to target, we find that it chooses the closest and second closest fires 15\% of the time, respectively, while choosing the third closest fire 68\% of the time. Since fires are arranged in clusters of three, and the separation between clusters is much larger than the separation between the fires within them, agents appear to choose the closest cluster of fires 98\% of the time. It is important to note that if an agent is currently at a fire, then moving to the closest fire is equivalent to staying where it is and immediately making a new decision. Though it is peculiar that the \textit{third} closest fire is being chosen with the highest probability, there is little cost to moving to another fire when already at a cluster. 

We encode a policy with the behaviors listed in the previous paragraph as follows. If the agent is at a live fire, it will hold with 84\% probability if there is no other interest. If there is other interest, it will hold with 27\% probability. If it does not hold, will fly to other fires with the probabilities listed above. A flowchart representing this policy is shown in \Cref{fig:appxpolicy}. When testing this simpler approximation of the learned policy, we find that it performs 28\% worse than the learned policy when compared on average discounted return. Thus, although this simpler policy seems to approximate the learned policy in important ways, there appear to be subtleties in the learned policy that improve performance. 

With confidence that policies training in this problem domain learn intelligent behaviors, we compare the effects of varying the fixed time-step of simulation on the policy learned. All policies were trained for 300 epochs, after which all training curves had reached asymptotes as policies settled to local optima. For each simulation environment, five separate policies were trained from scratch, to account for variation in the local optima on which they may settle. Each of these policies were then tested on the event-driven simulation environment, and the average discounted returns for each of policies were averaged to estimate the performance of a policy learned on that simulation environment and tested on the event-driven environment. 

\Cref{fig:transferperformance} shows performance statistics for fixed time-step simulation environments with time-steps $\Delta T$ ranging from $0.1$\si{\second} to $10.0$\si{\second}. We see that when the time-step is less than one second, the learned policies perform as well as those learned on the event-driven simulation. However, if the time-step is greater than one second, the performance is substantially degraded. Since the nominal value for $T_\text{hold}$ is $3.0\pm0.03$\si{\second}, and the value for $T_\text{health}$ is $3.0$\si{\second}, half of agents' attempts to extinguish fires would require them to make an additional attempt at doing so, since the fire would not have been extinguished in the single attempt. However, if the simulation time-step is greater than one second, the time held at a fire by any agent is rounded to a value greater than $3.0$\si{\second}, resulting in a policy learned expecting that any agents' attempt to extinguish a fire will result in it being extinguished. This expectation causes the performance of policies learned on simulations with these time-steps to degrade when tested on the event-driven simulator. 

To confirm that this is the cause of policy degradation, we modify the value of $T_\text{health}$ to be $2.9999$. Previously, the approximation made by a time-step of 1\si{\second} computed the fire to still be alive when it was supposed to be. However, with the modified value of $T_\text{health}$, this is no longer the case. That is, with this change, aircraft attempting to extinguish fires in the $\Delta T = 1$\si{\second} simulator will always see fires extinguished in a single attempt. As expected, this causes the performance of the policy to degrade when tested on the event-driven environment. 

This experiment provides an example of how the assumed time-step can force a policy to be learned that transfers poorly to an event-driven simulation, or the real-world, where temporal approximations are not made. While this experiment was tailored to demonstrate the problem, we can generally expect the choice of time-step to cause poor policy transfer if it is large enough to obfuscate the sequence in which events occur. In the Wildfire Fighting Problem, the performance degradation occurs because the order between the event of a hold-action ending and the fire being extinguished are obfuscated. If restricted to using fixed time-step simulation, we would thus hope to maintain a time-step that is much smaller than the typically duration that separates important events.

\subsection{Scaling the Time-Step to avoid Race-Conditions}
\label{sec:racecon}

Above, we defined a \textit{race condition} to be a time-step in which more than one event occurs. For an $N$-agent problem, we can assume that the time at which events pertaining to the $N$ agents occur are drawn from some distribution $p(t)$ for $t\geq0$. Hence, for $N$ samples from $p(t)$, a race condition occurs if any two samples fall into the same interval of size $\Delta T$. We can always shrink $\Delta T$ arbitrarily so that the probability of a race condition is less than some threshold $\delta$. However, since this comes at the cost of computation time, it is of interest to examine how $\Delta T$ must shrink as we increase $N$. We do so by selecting some example distribution, sampling 10,000 sets of $N$ samples from the distribution, and for each set of $N$ samples, we check for a race condition by seeing if more than two of the $N$ samples fall in the same interval of some chosen $\Delta T$. We then vary $\Delta T$ until the probability of a race condition is less than some threshold value of $\delta = 0.1$. To determine a relationship for each distribution tested, we vary $N$ from 2 to 20, solve for the corresponding $\Delta T$, and fit the functional form:
\begin{equation}
\Delta T = \alpha N^\beta
\end{equation}
\Cref{tab:nvdt} shows that for various example distributions, $\Delta T$ varies inversely with approximately $N^{2.2}$. For a fixed duration simulation, computation time scales inversely with the time-step. Hence, we find that computation time in fixed time-step simulations of event-driven process scales approximately as $N^{2.2}$ with the number of agents $N$, assuming we maintain a fixed low-probability of race conditions. Though the underlying distribution $p(t)$ of an arbitrary environment is unlikely to exactly be any of the example distributions chosen, this experiment suggests that an arbitrary distribution will also have $\beta \approx 2.2$, or at the very least, $\beta > 1$. Event-driven simulators, on the other hand, will process a number of events which are in many cases linearly related to the number of agents, providing a substantial computational advantage when simulating large-scale multi-agent problems.

\begin{table}
\centering
\caption{Relationship between $\Delta T$ and $N$, for maintaining constant $\delta$, as the timing of events are drawn from various distributions}
\label{tab:nvdt}
\begin{tabular}{@{}llr@{}} \toprule
Distribution & Parameters & $\beta~(\text{Std. Error})$ \\ \midrule
Exponential & $\lambda=1$ & $2.216~(0.055)$ \\ 
Chi-Squared & $k=5$ & $2.213~(0.045)$  \\
Gamma & $k=5, \theta=1$ & $2.219~(0.037)$ \\ 
\bottomrule
\end{tabular}
\end{table}

\section{Conclusion}

This paper presented an algorithm for learning neural network policies for control in cooperative multi-agent environments with asynchronous and temporally extended actions. The novelty of this contribution is that it extends an algorithm called PS-TRPO that does not require discretization of the observation space. This contrasts with existing algorithms tackling the same class of problem. \Cref{sec:bbcontrol} showed that our algorithm is able to learn an optimal policy for real-time bus holding, and is able to achieve better qualitative and quantitative performance than a sensible baseline policy. 

The approach to extending PS-TRPO involves framing the decentralized multi-agent decision problem as event-driven, allowing us to in many circumstances model our environment using event-driven simulation. Using such a simulator does not require assuming a fixed time-step, allowing us to eliminate \textit{race conditions}. \Cref{sec:feprob} showed that these artifacts from temporal discretization obfuscated the event sequences and can result in learning policies that are sensitive to the time-step. Further, \Cref{sec:racecon} showed that arbitrarily shrinking the time-step to avoid these artifacts scales poorly with the number of agents, motivating the use of event-driven simulation. The source code for this work can be found at \url{https://github.com/sisl/event-driven-rllab}. The algorithm presented in this paper is built as a modification to the TRPO framework in \texttt{rllab}~\cite{duan2016benchmarking}.

The work presented here borrows inspiration from hierarchical reinforcement learning, which attempts to simplify learning problems by stratifying decisions in levels of abstraction. However, the problem of simultaneously optimizing high and low-level policies is an open area of research. Our framework has assumed that we are only attempting to optimize a high-level policy, but the logical next step is to relax this assumption.

\appendices

\section*{Acknowledgment}

This work is based on work supported by NASA Grant No. NNXI4AIlIG and ARO Grant No. W911NF-15-1-0127. The authors acknowledge the support of Guillaume P. Brat from the NASA Ames Research Center, as well as the patient help from Jayesh K. Gupta. The authors would also like to thank the anonymous reviewers for their helpful comments.

\ifCLASSOPTIONcaptionsoff
  \newpage
\fi

{\small
\bibliographystyle{IEEEtran}

\bibliography{thebib}

\begin{thebibliography}{10}
\providecommand{\url}[1]{#1}
\csname url@samestyle\endcsname
\providecommand{\newblock}{\relax}
\providecommand{\bibinfo}[2]{#2}
\providecommand{\BIBentrySTDinterwordspacing}{\spaceskip=0pt\relax}
\providecommand{\BIBentryALTinterwordstretchfactor}{4}
\providecommand{\BIBentryALTinterwordspacing}{\spaceskip=\fontdimen2\font plus
\BIBentryALTinterwordstretchfactor\fontdimen3\font minus
  \fontdimen4\font\relax}
\providecommand{\BIBforeignlanguage}[2]{{%
\expandafter\ifx\csname l@#1\endcsname\relax
\typeout{** WARNING: IEEEtran.bst: No hyphenation pattern has been}%
\typeout{** loaded for the language `#1'. Using the pattern for}%
\typeout{** the default language instead.}%
\else
\language=\csname l@#1\endcsname
\fi
#2}}
\providecommand{\BIBdecl}{\relax}
\BIBdecl

\bibitem{sutton1999between}
R.~S. Sutton, D.~Precup, and S.~Singh, ``Between {MDP}s and {S}emi-{MDP}s: A
  framework for temporal abstraction in reinforcement learning,''
  \emph{{A}rtificial {I}ntelligence}, vol. 112, no. 1-2, pp. 181--211, 1999.

\bibitem{ghavamzadeh2006hierarchical}
M.~Ghavamzadeh, S.~Mahadevan, and R.~Makar, ``Hierarchical multi-agent
  reinforcement learning,'' \emph{International Conference on Autonomous Agents
  and Multiagent Systems (AAMAS)}, vol.~13, no.~2, pp. 197--229, 2006.

\bibitem{liu2016learning}
M.~Liu, C.~Amato, E.~P. Anesta, J.~D. Griffith, and J.~P. How, ``Learning for
  decentralized control of multiagent systems in large, partially-observable
  stochastic environments.'' in \emph{AAAI Conference on Artificial
  Intelligence (AAAI)}, 2016, pp. 2523--2529.

\bibitem{shen2006multi}
J.~Shen, G.~Gu, and H.~Liu, ``Multi-agent hierarchical reinforcement learning
  by integrating options into {MAXQ},'' in \emph{{I}nternational
  {M}ulti-{S}ymposiums on {C}omputer and {C}omputational {S}ciences
  ({IMSCCS})}, vol.~1.\hskip 1em plus 0.5em minus 0.4em\relax IEEE, 2006, pp.
  676--682.

\bibitem{clement2016temporal}
D.~M. Clement and M.~Huber, ``Temporal and agent abstractions in multiagent
  reinforcement learning,'' in \emph{IEEE International Conference on Systems,
  Man, and Cybernetics ({SMC})}, 2016, pp. 002\,190--002\,195.

\bibitem{guptacooperative}
J.~K. Gupta, M.~Egorov, and M.~J. Kochenderfer, ``Cooperative multi-agent
  control using deep reinforcement learning,'' in \emph{{A}daptive {L}earning
  {A}gents {W}orkshop (ALA)}, 2017.

\bibitem{schulman2015high}
J.~Schulman, P.~Moritz, S.~Levine, M.~Jordan, and P.~Abbeel, ``High-dimensional
  continuous control using generalized advantage estimation,''
  \emph{International Conference on Learning Representations (ICLR)}, 2016.

\bibitem{schulman2015trust}
J.~Schulman, S.~Levine, P.~Abbeel, M.~Jordan, and P.~Moritz, ``Trust region
  policy optimization,'' in \emph{International Conference on Machine Learning
  (ICML)}, 2015, pp. 1889--1897.

\bibitem{amato2014planning}
C.~Amato, G.~D. Konidaris, and L.~P. Kaelbling, ``Planning with macro-actions
  in decentralized {POMDP}s,'' in \emph{International Conference on Autonomous
  Agents and Multiagent Systems (AAMAS)}, 2014, pp. 1273--1280.

\bibitem{simpy}
\BIBentryALTinterwordspacing
K.~M\"{u}ller and T.~Vignaux, ``{SimPy: Simulating Systems in Python},''
  2002--. [Online]. Available: \url{https://pypi.python.org/pypi/simpy}
\BIBentrySTDinterwordspacing

\bibitem{chen2016real}
W.~Chen, K.~Zhou, and C.~Chen, ``Real-time bus holding control on a transit
  corridor based on multi-agent reinforcement learning,'' in \emph{IEEE
  International Conference on Intelligent Transportation Systems ({ITSC})},
  2016, pp. 100--106.

\bibitem{storn1997differential}
R.~Storn and K.~Price, ``Differential evolution--a simple and efficient
  heuristic for global optimization over continuous spaces,'' \emph{Journal of
  Global Optimization}, vol.~11, no.~4, pp. 341--359, 1997.

\bibitem{scipy}
\BIBentryALTinterwordspacing
E.~Jones, T.~Oliphant, P.~Peterson \emph{et~al.}, ``{SciPy}: Open source
  scientific tools for {Python},'' 2001--. [Online]. Available:
  \url{http://www.scipy.org/}
\BIBentrySTDinterwordspacing

\bibitem{duan2016benchmarking}
Y.~Duan, X.~Chen, R.~Houthooft, J.~Schulman, and P.~Abbeel, ``Benchmarking deep
  reinforcement learning for continuous control,'' in \emph{International
  Conference on Machine Learning (ICML)}, 2016.

\end{thebibliography}
}

\end{document}